\documentclass[conference]{IEEEtran}
\IEEEoverridecommandlockouts
\usepackage{cite}
\usepackage{amsmath,amssymb,amsfonts}
\usepackage{graphicx}
\usepackage{textcomp}
\usepackage[dvipsnames]{xcolor}
\usepackage{caption}
\usepackage{algorithm}
\usepackage{algpseudocode}
\usepackage{enumitem}
\usepackage{dblfloatfix}
\usepackage[nonumberlist,acronym,nomain]{glossaries}
\newacronym{cbp}{CBP}{Compute-Bandwidth-Precision}
\newacronym{sc}{SC}{Split Computing}
\newacronym{mse}{MSE}{Mean Square Error}
\newacronym{miou}{mIoU}{mean Intersection over Union}
\newacronym{rtt}{RTT}{Round Trip Time}
\newacronym{sota}{SotA}{State-of-the-art}
\newacronym{map}{mAP}{mean Average Precision}
\newacronym{rpi4}{RPi4}{Raspberry Pi 4}

\def\BibTeX{{\rm B\kern-.05em{\sc i\kern-.025em b}\kern-.08em
    T\kern-.1667em\lower.7ex\hbox{E}\kern-.125emX}}

\begin{document}

\title{Slimmable Encoders for Flexible Split DNNs in Bandwidth and Resource Constrained IoT Systems
}

\author{\IEEEauthorblockN{Juliano S. Assine$^\dagger$, J. C. S. Santos Filho$^*$, Eduardo Valle$^*$ and  Marco Levorato$^\dagger$}

\IEEEauthorblockA{
\small{$^\dagger$Department of Computer Science, University of California, Irvine, United States} \\
\small{$^*$School of Electrical and Computing Engineering (FEEC), Unicamp, Campinas, Brazil}\\
\small{E-mails: $\{$jsilotoa, levorato$\}$@uci.edu, $\{$jcssf, evalle$\}$@unicamp.br}
}
}



\maketitle

\begin{abstract}
The execution of large deep neural networks (DNN) at mobile edge devices requires considerable consumption of critical resources,  such as energy, while imposing demands on hardware capabilities. In approaches based on edge computing the execution of the models is offloaded to a compute-capable device positioned at the edge of 5G infrastructures. The main issue of the  latter class of approaches is the need to transport information-rich signals over wireless links with limited and time-varying capacity. The recent split computing paradigm attempts to resolve this impasse by distributing the execution of DNN models across the layers of the systems to reduce the amount of data to be transmitted while imposing minimal computing load on mobile devices. In this context, we propose a novel split computing approach based on slimmable ensemble encoders. The key advantage of our design is the ability to adapt computational load and transmitted data size in real-time with minimal overhead and time. This is in contrast with existing approaches, where the same adaptation requires costly context switching and model loading. Moreover, our model outperforms existing solutions in terms of compression efficacy and execution time, especially in the context of weak mobile devices. We present a comprehensive comparison with the most advanced split computing solutions, as well as an experimental evaluation on GPU-less devices.   \end{abstract}

\begin{IEEEkeywords}
Edge Computing, Split Deep Neural Networks, Internet of Things, Slimmable Encoders.
\end{IEEEkeywords}

\section{Introduction}
Bringing the full power of deep learning to edge and mobile devices requires overcoming two critical resource constraints: computing power and channel bandwidth. If the algorithms are executed locally at the mobile devices, then the main issue is the weak computing power and small energy reservoir of this class of devices, which likely results in low performance and/or high latency and limited lifetime.
If the execution of the algorithms is offloaded to infrastructure-level devices -- e.g., edge servers, then the limited and time-varying capacity of the communication channels connecting the mobile/edge devices to the servers may result in large latency and latency variance, while also posing considerable
infrastructure-level resource consumption (e.g., channel capacity and server time). The two options induce an intuitive tradeoff on resource usage at different layers of the overall system, as well as on crucial performance metrics. Recently, further expanding the array of computing options and thus operating points in the tradeoff, the \gls{sc}~\cite{matsubara2021split} paradigm emerged as an important research trend.

In \gls{sc}, deep neural networks (DNN) are split into two sections, executed on the mobile/edge device and an edge server, respectively. The ultimate objective is to balance energy consumption, channel usage, task performance, and overall latency. Early approaches use the original architecture and weights \cite{kang2017neurosurgeon, xu2019deepwear}, and thus the inherent ``compression'' capabilities of DNN architectures to cast optimization problems focused on the tradeoff between computing load at the mobile device, channel usage, and latency. In more recent \gls{sc} approaches~\cite{matsubara2021split}, the architecture and training of the DNNs are altered to boost the ability of the network to compress information in the early layers, thus improving the operating point in the compute/communication tradeoff. Importantly, this latter class of models often exploits the fact that the output representation of the compression portion of the altered model is not meant to reconstruct the original input but to support the computing task itself. Notably, in computer vision tasks this also results in improved privacy provided by the system.
\begin{figure}[tbp]
\centerline{\includegraphics[width=0.99\linewidth]{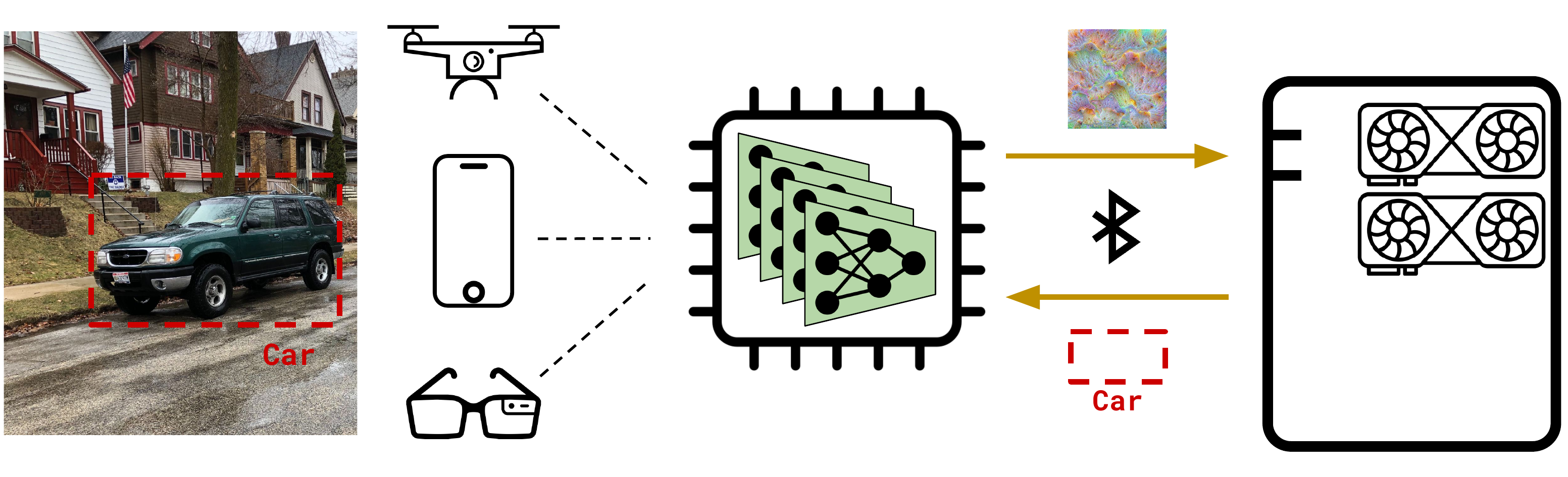}}
\caption{\textbf{Overall schematics of the proposed system}, where slimmable encoders are trained to produce adaptive compression toward an object detection task.}
\label{fig:application}
\end{figure}

\gls{sc} approaches can produce more than an order of magnitude compression ratios compared to classical approaches such as JPEG, thus enabling the support of complex DNN-based analysis in systems with constrained channel capacity. Nonetheless, the practicality of \gls{sc} in real-world settings remains to be proven, especially due to two main shortcomings of existing \gls{sc} frameworks:

\vspace{1.5mm}
\noindent
{\bf 1} - \emph{No dedicated hardware}: unlike JPEG/MPEG codecs, most devices either do not possess hardware dedicated to executing DNNs, or are equipped with GPUs which are likely shared with graphical tasks necessary to core system functions. As a consequence, the portion of the DNN executed at the edge device needs to have extremely low complexity.

\vspace{1.5mm}
\noindent
{\bf 2} - \emph{Heterogeneous Hardware and Operational Conditions}: Most \gls{sc} solutions need to be designed to match the specific characteristics of the edge device, and offer limited portability as well as limited ability to adapt to temporal variations. For instance, the complexity of the network portion executed at the mobile device is typically modified by changing the splitting point, but this maps to harsh tradeoffs with channel usage. Moreover, switching from one DNN model to another comes at the price of perceivable loading time of the model into the GPU memory.
In this work, we tackle these challenges by proposing an innovative \gls{sc} architecture that provides low-overhead low-latency online adaptability and is suited for CPU-only devices. 
At a high level, by altering the model architecture, advanced \gls{sc} techniques introduce in the model two sections corresponding to a neural encoder and decoder. The encoder portion attempts to produce a compact representation used by the decoder and tail portions to complete the original model's task.

Our proposed strategy results in a minimalist and highly adaptive encoder design. The proposed base encoder utilizes only 76M Flops and 1.2M activations, and the complexity can be increased or decreased at runtime to match computational needs and desired accuracy. This capability is granted by the proposed slimmable ensemble technique. Furthermore, bandwidth usage is also tunable by using flexible quantization levels based on the ensemble size and target output data size. By combining these techniques, the encoder portion of the model can be adapted at runtime both from the point of view of computing load and channel usage independently, providing the needed flexibility to adapt to channel state variations and device general capabilities and current state (e.g., residual energy). This training technique is not only novel in the use of slimmable training for ensembles, which has its own particularities in terms of aggregation, but also opens the possibility of slimmable models computed at different devices since each individual encoder is independent of the other.

Although the principles behind our approach are general, we provide a full implementation of our design for an object detection task. Specifically, we modify the architecture of the state-of-the-art model EfficientDet-D2\cite{tan2020efficientdet}. We compare its performance and characteristics with the main competitors available in the literature and demonstrate that our architecture is the only one enabling run-time adaptation on commodity hardware while providing up to 4x reduction in encoder latency and reducing memory usage by up to 2.8. Overall, our design has very high configurability with 16 possible modes of operation. Finally, we show experiments on a proof-of-concept system based on the Bluetooth 4.0 communication technology and demonstrate that our model can achieve near-real-time (${\sim} 200$ms) end-to-end latency. In this context, we illustrate its ability to adapt computational load and bandwidth usage as the channel characteristics shift due to device mobility in an indoor setting -- with up to 9m of distance between the mobile device and the edge server and data rate variability between 80kB/s and 200kB/s. To the extent of our knowledge, this is the first attempt to quantify the dynamic behavior of an \gls{sc} framework. 

The organization of the paper is as follows. Section \ref{sec:literature_review} provides a discussion of prior work, with a particular emphasis on recent \gls{sc} approaches that achieve state-of-the-art performance. These approaches will serve as the baseline for our experimental evaluation. In Section \ref{sec:direct_inverse}, we present an analysis of the tradeoffs commonly addressed in \gls{sc} framework design, and specifically we discuss how the relationship between the encoder, decoder, and bottleneck sizes affects design choices. In the remaining of Section \ref{sec:methods}, we present our framework and model, describing in detail the model architecture and the proposed novel slimmable ensemble technique. Section \ref{sec:experiments} presents results assessing the performance of our design and is organized into three parts. First, we validate our proposed training strategy by decoupling its components. Then, we perform an on-device comparison of the encoder performance with respect to the models in the literature with the highest performance. To this aim, we use a \gls{rpi4} board as a reference, showing that our solution has the best performance even on weak GPU-less devices. Finally, we present a full-system evaluation of the overall \gls{rtt} under static and dynamic network conditions. Section~\ref{sec:concl} concludes the paper.

\section{Literature Review}
\label{sec:literature_review}
In this section, we position our work within the SC area and provide background on critical components of our solution.

\subsection{Split Computing}
First, we provide a review of key concepts  and results related to SC. For a more comprehensive review of past work and future challenges, we refer the reader to \cite{matsubara2021split, bajic2021collaborative}.

Early attempts at splitting neural networks focused on off-the-shelf models to demonstrate the feasibility of the proposed techniques~\cite{kang2017neurosurgeon, xu2019deepwear}. In contrast, current \gls{sc} frameworks feature partially or entirely custom architectures with a bottleneck section meant to reduce the dimensionality of the representation at the split and use discretization/quantization of tensors to reduce bitrate~\cite{eshratifar2019bottlenet, matsubara2019distilled, choi2020back}. 

Most work so far, including this paper, resorted to a transform-coding paradigm common in traditional image compression. In fact, borrowing modern techniques from deep image compression often results in excessively complex encoders that require desktop-level GPUs to be executed and whose deployment on mobile devices is unfeasible~\cite{mentzer2019practical}. Moreover, the widespread use of costly floating-point generalized divisive normalization (GDN) activation functions~\cite{balle2015density} is detrimental to constrained devices prevalent in the \gls{sc} community. However, we note that there have been recent attempts at importing modern entropy model approaches into \gls{sc} techniques~\cite{matsubara2022supervised, matsubara2022sc2}, and in this paper, we use uniform noise regularization to improve quantization, as made popular by \cite{balle2016end}.

In this work, we focus on object detection, more specifically on the COCO2017 dataset~\cite{lin2014microsoft} - the most used dataset in the object detection literature. We report our findings for the \gls{map}@50:95 metric, obtained by averaging the \gls{miou} precision under the thresholds between 50\% and 95\%  with 5\% steps. We chose this setting to ensure a fair and uniform comparison, as it is the most used in the context of \gls{sc}~\cite{assine2021collaborative}. However, we note that other contributions used variations of the COCO dataset~\cite{choi2020back, cohen2020lightweight}.

We identify three main works that are the most related to our contribution. First, we consider the framework in~\cite{matsubara2022supervised}, whose \gls{sc} design makes use of an entropy model~\cite{Balle2018} embedded into a RetinaNet (with a ResNet backbone) baseline architecture. The design is validated on constrained devices such as \gls{rpi4}. In~\cite{assine2021collaborative}, the authors propose the first flexible architecture for \gls{sc} with scalable encoders. Different from this work, the solution is based on standard channel reduction techniques. Finally, Lee et al~\cite{lee2021splittable} proposed an encoder with remarkable efficiency by leveraging an aggressive 1-bit quantization of the bottleneck representation on a YoloV5 baseline architecture. The main proposition is a concurrent encoder downscaling with asymmetric decoder upscaling, resulting in almost no drop in accuracy when reducing the size of the compressed representation. The technique is demonstrated on multiple split points and validated in devices of the NVIDIA Jetson family. 

In addition to the conceptual innovation contained in our work, we show that our proposed approach significantly improves performance compared to existing solutions. To this aim, we have re-implemented the encoder architecture of the competing solutions (Fig.~\ref{fig:sota}) and obtained results detailed in Section \ref{sec:4:mobile}.

\subsection{Flexible Split Computing}

 The design of neural networks that can be reconfigured at runtime to match different operational conditions is a rather recent trend in the field of deep learning. One of the main lines of contributions proposes the development of slimmable neural networks~\cite{yu2018slimmable}, which provides a tool to navigate the precision-computation tradeoff for convolutional neural networks. This technique has been further advanced in~\cite{yu2019universally} and later given a formalized treatment in the context of network subspaces in~\cite{wortsman2021learning, nunez2021lcs}.
 
 We note that the use of slimmable techniques in split computing was first proposed in \cite{assine2021collaborative}. However, our design realizes robust improvements providing independent configurations of bandwidth usage and computing load, and achieves orders of magnitude better efficiency (as shown in Section \ref{sec:4:mobile}).

\section{Flexible Encoding in SC}
\label{sec:methods}
In this section, we describe the framework enabling flexible encoding in the context of SC. Our solution can easily transition across operating points in the trade-off between bandwidth usage and computing load. First, we describe the tradeoff, then, we present a general formulation of the concept and the specific implementation that is used in our experiments.

\subsection{Tradeoffs In Split Computing}
\label{sec:direct_inverse}
\begin{figure}[tpb]
\begin{center}
\includegraphics[width=0.95\linewidth]{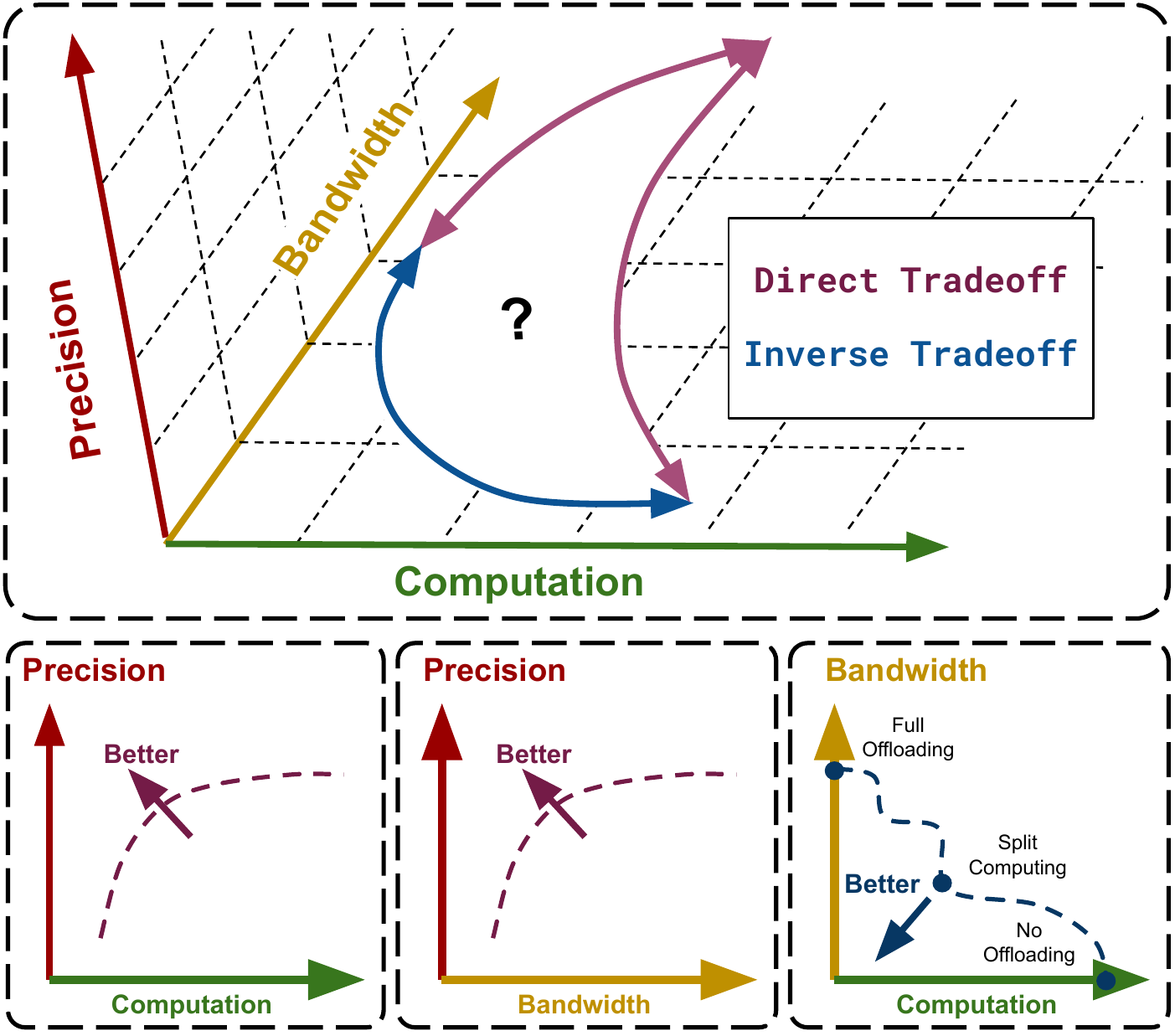}
   \caption{\textbf{The Direct-Inverse Tradeoff} is widely used in the literature but explored on a trial-and-error basis. Computing power on both the encoder and decoder, as well as bandwidth, are empirically known to have a monotonic increasing relationship with precision. Changing both of these resources at once however, such as in changing split points, is expected to have a monotonic decreasing relationship when precision is kept fixed, but the overall behavior of this tradeoff is still unclear.}
   \label{fig:dirinv_tradeoff}
\end{center}
\end{figure}

As pointed out in the previous sections, compression plays a central role in \gls{sc} design. In this section, we emphasize how SC generates tradeoffs that include other important aspects, such as computational costs. One of the main motivations to use \gls{sc} is to enable DNN models in settings where edge devices do not have enough computational power to perform a given inference task at all or to perform it in a feasible time. 

In SC, both compression and inference are performed in an end-to-end manner, so that the restrictions in terms of bandwidth and computing are almost indistinguishable. This becomes more clear using the example of a simple task such as image classification. In classification, the highest achievable compression rate is obtained when the whole model is executed locally by the edge device, which then transmits the few bits indicating the resulting class prediction. Intuitively, this approach collapses to solutions based solely on local computing. Most image classification architectures are built in such a way that successively smaller layers are used. Thus, many \gls{sc} approaches simply choose the latest feasible bottleneck that can be computed by the mobile device and alter it to reduce the number of nodes. Thus, split computing is usually portrayed as a task-oriented form of compression under a computational budget. 

In this paper, we frame the problem as a 3-dimensional tradeoff: namely the \gls{cbp} tradeoff induced in the encoder design, best understood visually with the help of Fig. \ref{fig:dirinv_tradeoff}. We note preliminary attempts in this direction by ~\cite{assine2019compressing}; and~\cite{shao2020communication}, where a technique for optimizing all three dimensions simultaneously is presented, but here we expect not only to describe its nuances further but also to allow the users to obtain the best \gls{cbp} operating point for the specific context with a single model.

The components of the \gls{cbp} tradeoff can hardly be analyzed separately due to their interdependence. For instance, when the size of the encoder is fixed, resizing the decoder changes the operating point~\cite{lee2021splittable} and vice-versa~\cite{assine2021collaborative}. If we fix the overall architecture, but change the splitting point, we also obtain non-trivial trends, as compression usually entails the addition of a reduction component exogenous to the model itself, which invalidates the assumption of a fixed architecture. A possible solution is to use the ratio between encoder and decoder as feature~\cite{dong2022splitnets}. The key advantage is that the tradeoff between computation and precision on the overall model is often monotonic with respect to this ratio. 

The motivation for split computing mostly originates from the restrictions on the encoder (e.g., number of parameters and output size). As a result, most SC solutions focus on the design of the encoder. In this context, we identify two main tradeoffs. The \textbf{Direct Tradeoff} describes the monotonic increase of resource usage as we increase the output precision. This setup includes the present work and prior contributions such as \cite{assine2021collaborative, matsubara2019distilled, matsubara2022supervised}. In the direct tradeoff, both computation and bandwidth constraints limit the model's expressiveness with a direct impact on precision. 
More widely explored, though least understood, is the \textbf{Inverse Tradeoff}, where the ratio between encoder and decoder is controlled, typically by splitting a primer architecture at a different location~\cite{assine2019compressing, eshratifar2019bottlenet, shao2020bottlenet++,  jankowski2020joint, lee2021splittable}. In this setting, a target precision is fixed while trading a larger computing load for a smaller encoder output's size. However, it is unclear whether the resulting trend is monotonic or continuous even in the architecture space.

\subsection{Model Design}
\label{sec:methods:framework}

First, we lay the ground for our \gls{sc} settings. Let's assume an existing architecture is split into two sections: an encoder $z = f'(x)$ and a decoder $\hat{y} = g'(z)$, where $x$ is the input, $z'$ is the intermediary representation and $\hat{y}'$ is the inference result. As explained earlier, our goal is to achieve a desirable size of the representation $z'$ -- under a complexity budget. This requires the structural modification of both the encoder and decoder, which we call $z = f(x)$ and $\hat{y} = g(z)$ respectively. Then, a quantization/dequantization stage is applied to the intermediary representation to further reduce the representation size. We denote these operations with $z_Q=Q(z)$, $\Tilde{z}=Q^{-1}(z_Q)$, as the whole \gls{sc} sequence can be expressed as 
\begin{equation}
\hat{y}=g\circ Q^{-1} \circ Q \circ f(x).
\end{equation}

In this context, our solution adopts a distillation training approach inspired by that proposed in~\cite{matsubara2020head}, where the network is trained using a \gls{mse} loss between the original (teacher) and modified (student) architectures. Both the teacher's decoder $g'$ and the student's decoder $g$ are further split into trainable and frozen sections $g'=g'_t \circ g'_f$ and $g=g_t \circ g_f$. Notably, while at deployment time we need to compute $\hat{y}$, during training the models are only executed up to $r=g_t(\Tilde{z})$ and $r'=g'_t(z)$. The training loss, then, is:
\begin{equation}
\mathcal{L}_{mse}= \frac{||r-r'||^2}{dim(r)}.
\end{equation}

This formulation has the advantage of faster convergence as there is no need to train the whole model from scratch, and, most importantly, results in a self-supervised form of training, meaning that training does not require the labels.

One of the key novelties of our design is the use of an ensemble of $N$ encoders $f_i(x)$, $i \in [1 \dots N]$ as a drop-in replacement for the regular encoder. At any inference round, the ensemble can be subsampled in an ordered fashion, which is equivalent to selecting the index (size) $s \in [1 \dots N]$. The output of an encoder of size s is then calculated as:
\begin{equation}
f_s(x) = \sum_{i=1}^s \frac{1}{2^{i-1}}f_i(x).
\end{equation}
We will explain in detail the architecture of the ensemble encoders in the next section, where we focus on a specific task and model.

In general, removing members of a jointly trained ensemble could cause unpredictable performance degradation. We prevent this effect by introducing a slimmable training technique to achieve a monotonic tradeoff between the ensemble size and precision, as well as by normalizing each output $f_i(x)$ such that the range of $z=f(x)$ is predictable, improving training stability. At each step of training, being $x$ the training sample and $\mathcal{D}$ the dataset, 
we sample a set of $S$ sizes. This set always contains the smallest and largest sizes, that is, $s=1$ and $s=N$, and a choice of $S-2$ random sizes in $s \in [1 \dots N]$. The fixed choice of the smallest and largest sizes is known as the \textit{"sandwich rule"}~\cite{yu2019universally}, which improves the quality of slimmable channel training and has been verified empirically to provide good performance in ensembles. A precise description of the training procedure is reported in Algorithm~\ref{alg:training} and depicted in Fig.~\ref{fig:training}.

\begin{algorithm}[t]
\begin{algorithmic}

\State $s_n \gets [1, \dots N]$
\State $Teacher \gets g_t' \circ f'$
\State $Student \gets g_t \circ Q^{-1} \circ Q \circ f_s$
\\

\For{ Sample($x$) in  Dataset($\mathcal{D}$)}
\State $l \gets 0$
\State $l \gets l + \mathcal{L}_{mse}(Student(1, x), Teacher(x))$
\State $l \gets l + \mathcal{L}_{mse}(Student(N, x), Teacher(x))$
\For{$i$ in $range(S-2)$ }
\State $l \gets l + \mathcal{L}_{mse}(Student(choice(s_n), x), Teacher(x))$
\EndFor
\EndFor

\end{algorithmic}
\caption{Ensemble Slimmable Training}
\label{alg:training}
\end{algorithm}

Furthermore, to abate the size of the transmitted data, we convert the natural 32-bit representation into a quantized representation. To this aim, we use a  quantization with variable bounds, that gracefully accommodates ensembles of different sizes. The quantization function $Q(z)$  and dequantization $Q^{-1}(z_Q)$ are described by the equations 
\begin{equation}
  z_Q = \left\lfloor{Clip \left((2^b-1)\left(\frac{z}{b*\sigma(z)} + 0.5\right), 0, 2^b-1\right)}\right\rceil,
  \label{eq:quantization1}
\end{equation}
\begin{equation}
  \Tilde{z}= \left(\frac{z_Q}{2^b-1}-0.5\right)*b*\sigma(z_Q),
  \label{eq:quantization2}
\end{equation}
 where $b$ is the bitrate per symbol and $\sigma$ is the standard deviation.

Finally, we employ a commonly used technique to improve robustness to uniform quantization: the addition of uniform noise during training. In our setting, we need to scale noise to match the size of the ensemble:
\begin{equation}
  \Tilde{z} = Q^{-1} \circ Q(z) = z+\mathcal{U}(-2^{-s}, 2^{-s}).
  \label{eq:regularization}
\end{equation}

\subsection{Model Implementation}
\label{sec:methods:implementation}

\noindent
\begin{figure}[]
\centering
\includegraphics[width=0.99\linewidth]{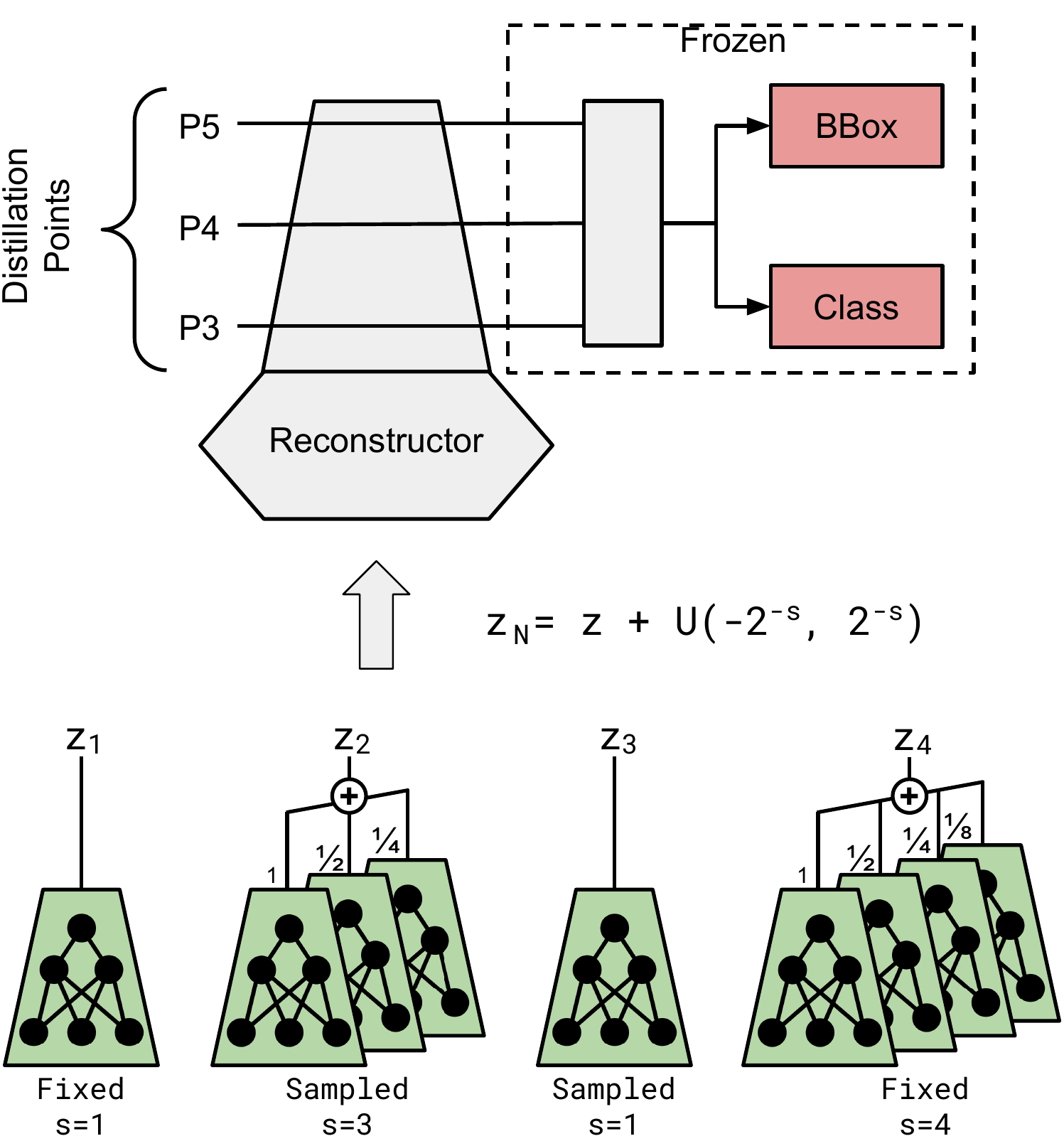}
\captionof{figure}{\textbf{Visualization of the slimmable training process.} At each step, multiple copies of the encoder with different sizes are replicated. Noise regularization is added between the encoder and decoder to improve robustness to quantization.}
\label{fig:training}
\end{figure}

\begin{figure}[t!]
  \centering
   \includegraphics[width=0.99\linewidth]{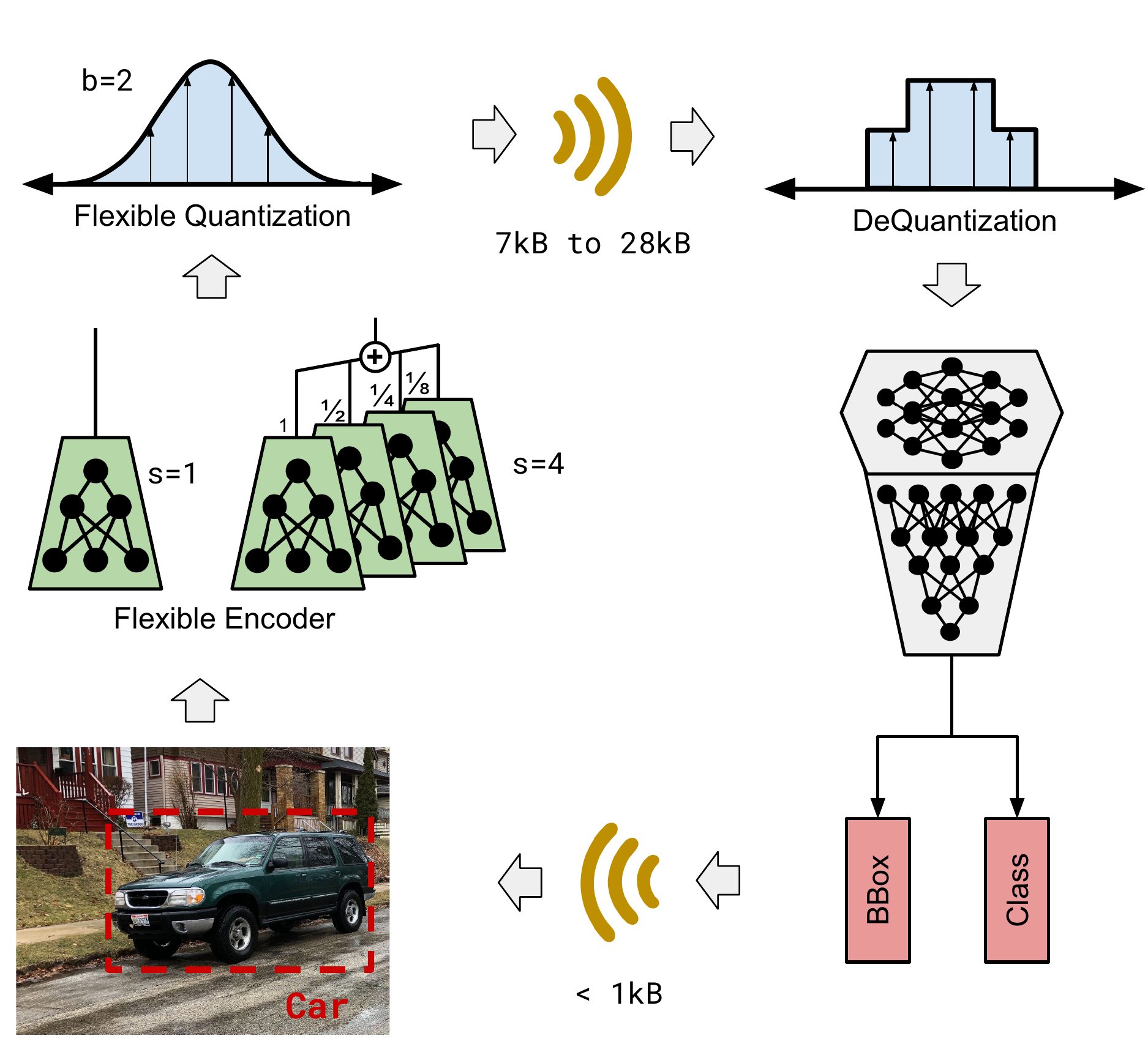}
   \caption{\textbf{Depiction of the Model Implementation} }
   \label{fig:system_framework}
\end{figure}

We now describe the implementation of the framework previously presented by focusing on an object detection task. We select as a baseline the EfficientDet(D2) architecture~\cite{tan2020efficientdet}. The original model\footnote{The reference implementation can be found in the repository github.com/zylo117/Yet-Another-EfficientDet-Pytorch} is split after the second bottleneck into encoder $f'$ and decoder $g'$, and is used as a teacher model by splitting $g'$ at the output points $r'=[P_3, P_4, P_5]$. Referring to common object detection nomenclature, we use the connections between the backbone and the neck~\cite{bochkovskiy2020yolov4} of the model as distillation points, as shown in Fig.~\ref{fig:training}.

\begin{figure}[h]
\centering
\includegraphics[width=0.5\linewidth]{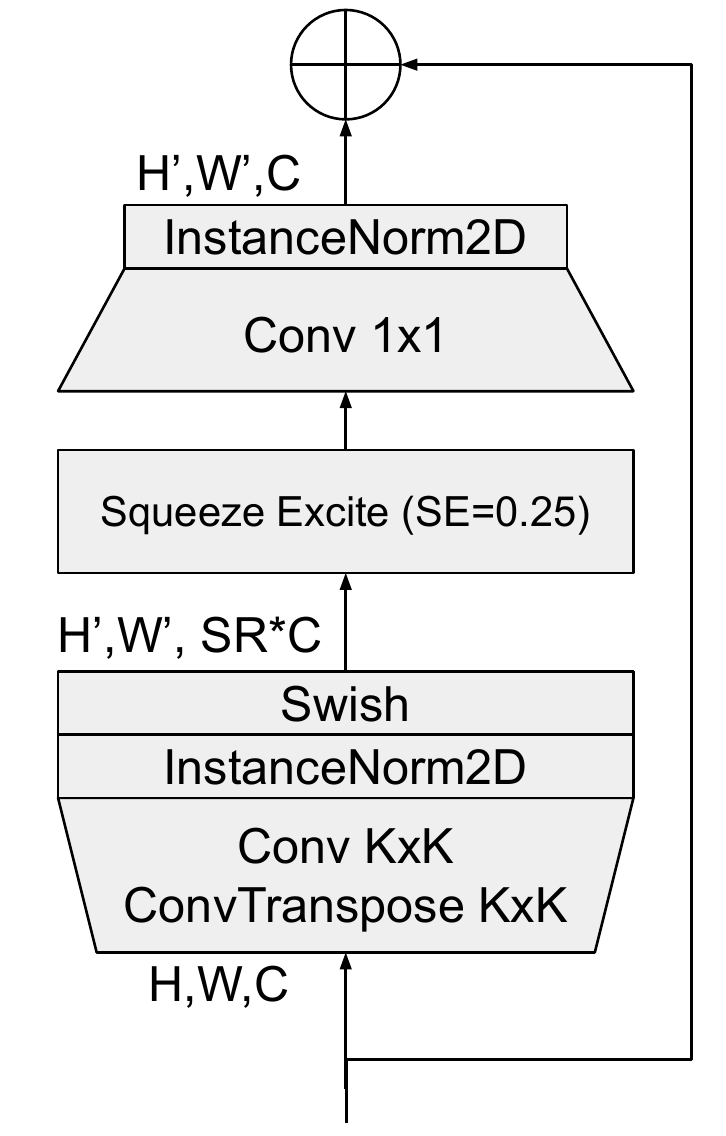}
\captionof{figure}{\textbf{Fused-MBConv} blocks use regular convolutions and \textbf{Fused-MBConvT} use transpose convolutions.}
\label{fig:convblock}
\end{figure}

In order to describe $f$ and $g$, we first define a basic building block, repeated across layers, used in all our custom modules: a modified version of the EfficientNetV2 block~\cite{tan2021efficientnetv2}. In our version, we use instance normalization instead of batch normalization, but for simplicity, we will keep the original nomenclature of Fused-MBConv when referring to it. Additionally, transpose convolutions are used when upscaling is required, which we will refer to as Fused-MBConvT. A visual representation of this module is provided in Fig.~\ref{fig:convblock}.

\noindent

\begin{table*}[b!]
  \centering
    \caption{\textbf{Ensemble Size}: Ensemble encoders are all trained with a maximum size of 4 and compared with a single encoder trained with no ensemble. One can see that in almost all cases the ensemble training beats the single encoder even when $s=1$. \\ \textbf{Quantization Performace}: We compare trainings with added uniform quantization as described in section \ref{sec:methods}. Overall results are improved by the noise regularization when quantization is applied.}
  \label{tab:ensemble_ablation}
  \resizebox{0.8\textwidth}{!}{
    \begin{tabular}{|l|llllllllll|}
    \hline
     & \multicolumn{10}{c|}{\textbf{mAP}} \\ \cline{2-11} 
    \textbf{Encoder} & \multicolumn{5}{c|}{\textbf{No Regularization}} & \multicolumn{5}{c|}{\textbf{Noise Regularization}}\\ \cline{2-11} 
     & \rule{0pt}{0.9\normalbaselineskip}fp32 & 4bits & 3 & 2 & \multicolumn{1}{l|}{1} & fp32 & 4bits & 3 & 2 & 1 \\ \hline
     
    \rule{0pt}{0.8\normalbaselineskip}Single & 37.2 & 34.7 & 30.5 & 20.0 & \multicolumn{1}{l|}{3.5} & 34.4 & 32.9 & 31.4 & 26.6 & 12.1 \\ \hline
    \rule{0pt}{0.9\normalbaselineskip}Ensemble(s=1) & 37.7 & 33.3 & 29.0 & 20.1 & \multicolumn{1}{l|}{5.3} & 35.7 & 34.3 & 32.0 & 27.7 & 14.5 \\
    Ensemble(s=2)& 37.9 & 35.0 & 31.3 & 23.0 & \multicolumn{1}{l|}{7.2} & 38.0 & 36.1 & 34.1 & 29.4 & 16.5 \\
    Ensemble(s=3) & 38.4 & 35.2 & 31.6 & 23.0 & \multicolumn{1}{l|}{7.1} & 38.3 & 36.4 & 34.2 & 29.4 & 16.4 \\
    Ensemble(s=4) & 38.5 & 35.5 & 31.8 & 23.2 & \multicolumn{1}{l|}{7.2} & 38.6 & 36.8 & 34.6 & 29.7 & 16.6 \\
    \hline
    \end{tabular}
    }
\end{table*}

 The original encoder is discarded in favor of a custom slimmable ensemble $f$ with maximum size $N=4$. Each individual encoder $f_i$ is an identical copy of the sequential architecture described in Table \ref{tab:encoder}, all outputs are normalized by the InstanceNorm layers and all inputs are equal but the size is downscaled from the original input size 768x768 to 384x384. Furthermore, the decoder is augmented with a reconstructor module $g_r$, that is $g = g' \circ g_r$. This reconstructor model, described in Table \ref{tab:reconstructor}, has a much larger size compared to the encoder. This choice makes more pronounced the asymmetric relationship between the encoder and decoder, which is necessary to achieve the desired precision while keeping the encoder size extremely small.

\begin{table}
\centering
\captionof{table}{\textbf{Encoder Architecture}}
\label{tab:encoder}
\scalebox{1.2}{ 
\begin{tabular}{|l|l|l|l|l|l|}
\hline
\textbf{Stage} & \textbf{Operator} & \textbf{C} & \textbf{S} & \textbf{K} & \textbf{Skip} \\ \hline
\multicolumn{1}{|l|}{1} & \multicolumn{1}{l|}{Conv2D} & \multicolumn{1}{|l|}{6} & \multicolumn{1}{l|}{2} & \multicolumn{1}{l|}{3} & - \\
\multicolumn{1}{|l|}{2} & \multicolumn{1}{l|}{Fused-MBConv, SR1} & \multicolumn{1}{l|}{4} & \multicolumn{1}{l|}{1} & \multicolumn{1}{l|}{3} & True \\
\multicolumn{1}{|l|}{3} & \multicolumn{1}{l|}{Fused-MBConv, SR1} & \multicolumn{1}{l|}{4} & \multicolumn{1}{l|}{1} & \multicolumn{1}{l|}{3} & True \\
\multicolumn{1}{|l|}{4} & \multicolumn{1}{l|}{Fused-MBConv, SR6} & \multicolumn{1}{l|}{6} & \multicolumn{1}{l|}{2} & \multicolumn{1}{l|}{3} & True \\
\multicolumn{1}{|l|}{5} & \multicolumn{1}{l|}{Fused-MBConv, SR6} & \multicolumn{1}{l|}{6} & \multicolumn{1}{l|}{1} & \multicolumn{1}{l|}{3} & True \\
\multicolumn{1}{|l|}{6} & \multicolumn{1}{l|}{Fused-MBConv, SR6} & \multicolumn{1}{l|}{6} & \multicolumn{1}{l|}{1} & \multicolumn{1}{l|}{3} & False \\ \hline
\end{tabular}
}

\centering
\captionof{table}{\textbf{Reconstructor Architecture}}
\label{tab:reconstructor}
\scalebox{1.2}{ 
\begin{tabular}{|l|l|l|l|l|l|}
\hline
\textbf{Stage} & \textbf{Operator} & \textbf{C} & \textbf{S} & \textbf{K} & \textbf{Skip} \\  \hline
\multicolumn{1}{|l|}{1} & \multicolumn{1}{l|}{Fused-MBConv, SR6} & \multicolumn{1}{l|}{48} & \multicolumn{1}{l|}{1} & \multicolumn{1}{l|}{1} & False \\
\multicolumn{1}{|l|}{2} & \multicolumn{1}{l|}{Fused-MBConv, SR6} & \multicolumn{1}{l|}{48} & \multicolumn{1}{l|}{1} & \multicolumn{1}{l|}{3} & True \\
\multicolumn{1}{|l|}{3} & \multicolumn{1}{l|}{Fused-MBConvT, SR6} & \multicolumn{1}{l|}{48} & \multicolumn{1}{l|}{2} & \multicolumn{1}{l|}{3} & False \\
\multicolumn{1}{|l|}{4} & \multicolumn{1}{l|}{Fused-MBConv, SR6} & \multicolumn{1}{l|}{48} & \multicolumn{1}{l|}{1} & \multicolumn{1}{l|}{3} & True \\
\multicolumn{1}{|l|}{6} & \multicolumn{1}{l|}{Fused-MBConv, SR6} & \multicolumn{1}{l|}{24} & \multicolumn{1}{l|}{1} & \multicolumn{1}{l|}{3} & False \\ \hline
\end{tabular}
}
\end{table}


\section{Experiments and Results}
\label{sec:experiments}
In this section, we assess the performance of the proposed solution and compare it thoroughly with the best-performing alternatives.

\subsection{Model Training}

First, we assess the proposed training technique (Section \ref{sec:methods:framework}) given the implementation described in Section \ref{sec:methods:implementation}. Training was performed using the ADAM optimizer over 20 epochs with a learning rate of 0.05, halving the learning rate every 5 epochs, all ensembles tested had maximum size $N=4$. Full results are presented in Table~\ref{tab:ensemble_ablation}.

We perform two forms of ablation to evaluate our technique, first, we compare a single encoder (i.e. $N=1$) with our ensemble encoder and show that even if when we set $s=1$ -- that is, a single active encoder -- our ensemble has a better performance compared to standard training (37.7\% vs 37.2\%). This outcome may result from an implicit form of regularization induced by training and motivates the use of our training technique even if only one encoder is incorporated in the architecture.

Subsequently, we explore the impact of quantization and regularization on the overall performance, each column in the table corresponds to one degree of numerical precision, from the original 32-bit floating point(fp32) up to a single bit of quantization(1). As expected, quantization results in a loss of precision, but, as we can see on the right-side panel in the table, the regularized versions outperform the non-regularized ones in all ensemble configurations, especially for more aggressive quantizations. Also, no degradation in performance is observed due to regularization, even for non-quantized results, in all but the $s=1$ setting.

Based on these results, we perform all subsequent experiments with the regularized version of our ensemble encoder, as quantization is often necessary in real-world settings.

\subsection{Mobile device Perfomance}
\label{sec:4:mobile}

We now evaluate the encoder's performance on target hardware compared to existing solutions. In this section, we restrict our experiments to the encoder portion of the model, which is often the most critical section due to the more stringent limitations of edge devices compared to edge servers. All experiments were performed on a \gls{rpi4} using the torchscript runtime environment\footnote{Operations not available in the torchscript environment, such as entropy coding were removed with no impact on the comparison.}.

\begin{figure*}[!t]
\begin{center}
\includegraphics[width=0.75\linewidth]{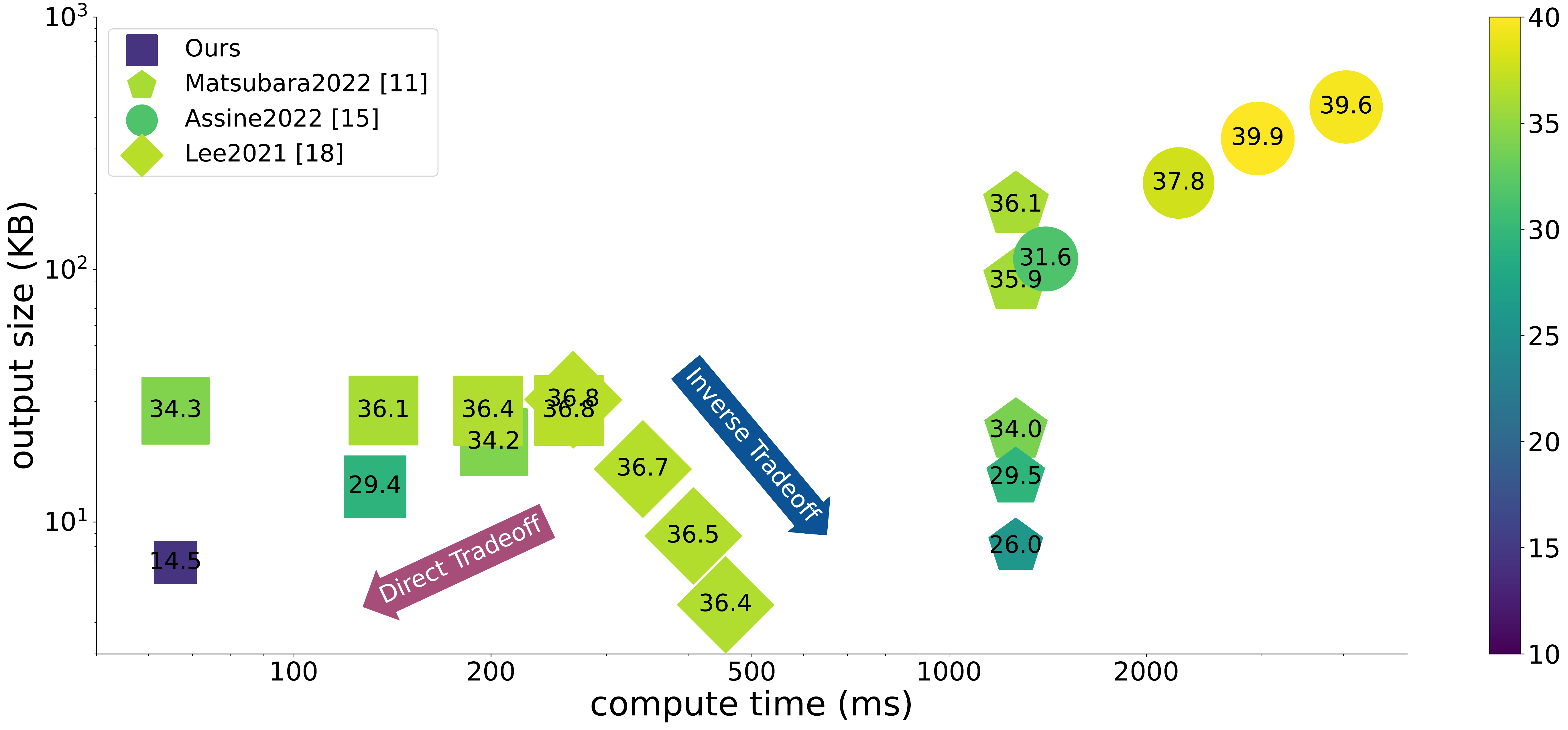}
\end{center}
   \caption{\textbf{Encoder Performance Comparison} between split object detection architectures on the \gls{cbp} axis. Inference times values were measured on a \gls{rpi4} device and are labeled and represented by size at each point.}
\label{fig:sota}
\end{figure*}

First, we analyze the impact of switching encoders at run-time. The results measuring RAM usage, disk load time, and warmup time are reported in Table~\ref{tab:switching}. We show that in order to switch between models at run-time, it is necessary to keep them in memory, as loading directly from disk and the subsequent model warmups require orders of magnitude larger time compared to inference time. However, the storage of multiple models in memory comes at the obvious price of memory usage, a precious resource, also due to the multifaceted role of computing resources in edge devices.
Importantly, our design provides the largest number of available configurations, while minimizing the memory necessary to load all configurations of the available models at the same time.

\begin{table}[h]
\caption{\textbf{Model Switching Performance:} Switching between model configs at runtime requires either re-loading each model from memory or keeping multiple models in memory resulting in aggregated memory usage}
\label{tab:switching}
\begin{tabular}{|c|c|c|c|c|}
\hline

\textbf{Model}          &     \textbf{Configs} &  \textbf{RAM(MB)} &  \textbf{Disk(ms)} &  \textbf{Warmup(ms)}  \\
\hline
Ours   &           16 &     85.7 &          954.0 &      1811.5 \\
Assine2022   &           12 &    204.9 &         1207.0 &      9913.6 \\
Matsubara2022 &            4 &    239.9 &          709.9 &     12785.1 \\
Lee2021       &            5 &    153.4 &         1264.5 &      5045.5 \\
\hline
\end{tabular}
\end{table}

Furthermore, we present a runtime comparison of all models on the \gls{cbp} metrics in Fig. \ref{fig:sota}. We note that such comparison may be limited as different models may follow different trajectories of the \gls{cbp} tradeoff. For instance, the two most competitive models, ours and  Lee2021\cite{lee2021splittable}, result in different trends. While we control both computation and output size directly impacting precision, their solution increases computing load in exchange for smaller compressed representations by designing models based on multiple split points. On the target device, our model is the only one capable of achieving a latency meeting typical requirements of real-time applications. For comparison, we also present statistics of all models in the picture in table \ref{tab:sota}, with common proxies for complexity such as the number of parameters, floating-point operations, and activations.

\begingroup
\setlength{\thickmuskip}{0mu}
\begin{table}
\caption{\textbf{Encoder Comparison:} Common statistics of computation performance such as Floating-Point Operations (FLOPs), number of parameters (Par) and activations (Act), as well as output size (Out) and \gls{map} (Mean Average Precision)}
\label{tab:sota}
\resizebox{\columnwidth}{!}{
\begin{tabular}{|l|c|c|c|c|c|}
\hline
\textbf{Model}  &    \textbf{FLOPs(M)} &  \textbf{Par(K)} &   \textbf{\gls{map}} &  \textbf{Out(KB)} & \textbf{Act(M)} \\
\hline
Ours (s=1, 1 bit)           &    75.7 &        6.3 &  14.5 &     6.9 &     1.2 \\

Ours (s=2, 2 bits)          &   151.4 &       12.7 &  29.4 &    13.8 &     2.5 \\
Ours (s=3, 3 bits)            &   227.1 &       19.0 &  34.2 &    20.7 &     3.7 \\
Ours (s=1, 4 bits)           &    75.7 &        6.3 &  34.3 &    27.6  &     1.2 \\
Ours (s=2, 4 bits)          &   151.4 &       12.7 &  36.1 &    27.6 &     2.5 \\

Ours (s=3, 4 bits)           &   227.1 &       19.0 &  36.4 &    27.6 &     3.7 \\
Ours (s=4, 4 bits)          &   302.8 &       25.3 &  36.8 &    27.6 &     5.0 \\
Lee2021 (layer 3)  &   541.5 &       29.8 &  36.8 &    30.5  &     5.3 \\
Lee2021 (layer 5)  &   913.0 &      143.3 &  36.7 &    16.2  &     7.6 \\
Lee2021 (layer 7)  &  1282.3 &      595.7 &  36.5 &     8.8  &     8.7 \\
Lee2021 (layer 10) &  1480.0 &     1089.8 &  36.4 &   4.7  & 9.2 \\
Matsubara2022 (1)  &  2897.1 &       63.3 &  36.1 &   180.0 &    11.5 \\
Matsubara2022 (2) &  2897.1 &       63.3 &  35.9 &    90.0 &    11.5 \\
Matsubara2022 (3)  &  2897.1 &       63.3 &  34.0 &    23.0 &    11.5 \\
Matsubara2022 (4)  &  2897.1 &       63.3 &  29.5 &    15.0 &    11.5 \\
Matsubara2022 (5)  &  2897.1 &       63.3 &  26.0 &     8.0 &    11.5 \\
Assine2022 ($\alpha=0.25$)   &   613.8 &       24.8 &  31.6 &   110.0 &    27.2 \\
Assine2022 ($\alpha=0.50$)  &  1083.4 &       60.1 &  37.8 &   220.0 &    42.5 \\
Assine2022 ($\alpha=0.75$)   &  1747.3 &      110.5 &  39.9 &   330.0  &    57.9 \\
Assine2022 ($\alpha=1.00$)  &  2605.4 &      176.9 &  39.6 &   440.0 &    73.3 \\
\hline
\end{tabular}
}
\end{table}
\endgroup

\subsection{System Evaluation}

In this section, we present an evaluation of an end-to-end system executing the proposed neural architecture. The system is comprised of a \gls{rpi4} acting as a sensor device, and a CPU server-side equipped with an Intel(R) Core(TM) i9-9820X and an NVIDIA GeForce RTX 2080 Ti GPU. The device and server are connected via Bluetooth 4.1. 

The objective of our experiments is to assess the round-trip time (RTT), from data acquisition to the availability of the task outcome. While pipelining on both server and device could lead to much higher frames per second providing an improved experience to the end user, the most important aspect for real-time applications is the total latency.

First, we present in Table \ref{tab:ideal} the results for all configurations on an optimistic scenario, where the edge device and server are in close proximity ($<$1m). For instance, this could correspond to a setting where both server and device are carried by the user (e.g., such as an augmented reality glasses and smartphone pair). It can be seen that in this setting near-real-time values of the \glspl{rtt} are achievable.

\begin{table}[tbp]
\caption{\textbf{RTT under Ideal System Conditions}}
\label{tab:ideal}
\begin{center}
\begin{tabular}{|c|c|c|c|}
\hline
\textbf{ms} & \textbf{\textit{\gls{map}}}& \textbf{\textit{Size}}& \textbf{\textit{Bits}} \\
\hline
211.2 &  14.5 &        1 &     1 \\
253.8 &  27.7 &        1 &     2 \\
289.9 &  16.5 &        2 &     1 \\
294.7 &  32.0 &        1 &     3 \\
330.8 &  34.3 &        1 &     4 \\
331.4 &  29.4 &        2 &     2 \\
361.7 &  16.4 &        3 &     1 \\
369.4 &  34.1 &        2 &     3 \\
399.8 &  29.4 &        3 &     2 \\
410.7 &  36.1 &        2 &     4 \\
438.2 &  16.6 &        4 &     1 \\
439.3 &  34.2 &        3 &     3 \\
476.8 &  29.7 &        4 &     2 \\
482.6 &  36.4 &        3 &     4 \\
515.3 &  34.6 &        4 &     3 \\
553.8 &  36.8 &        4 &     4 \\
\hline
\end{tabular}
\end{center}
\end{table}

\begin{figure}[b]
\centerline{
\includegraphics[width=0.7\linewidth]{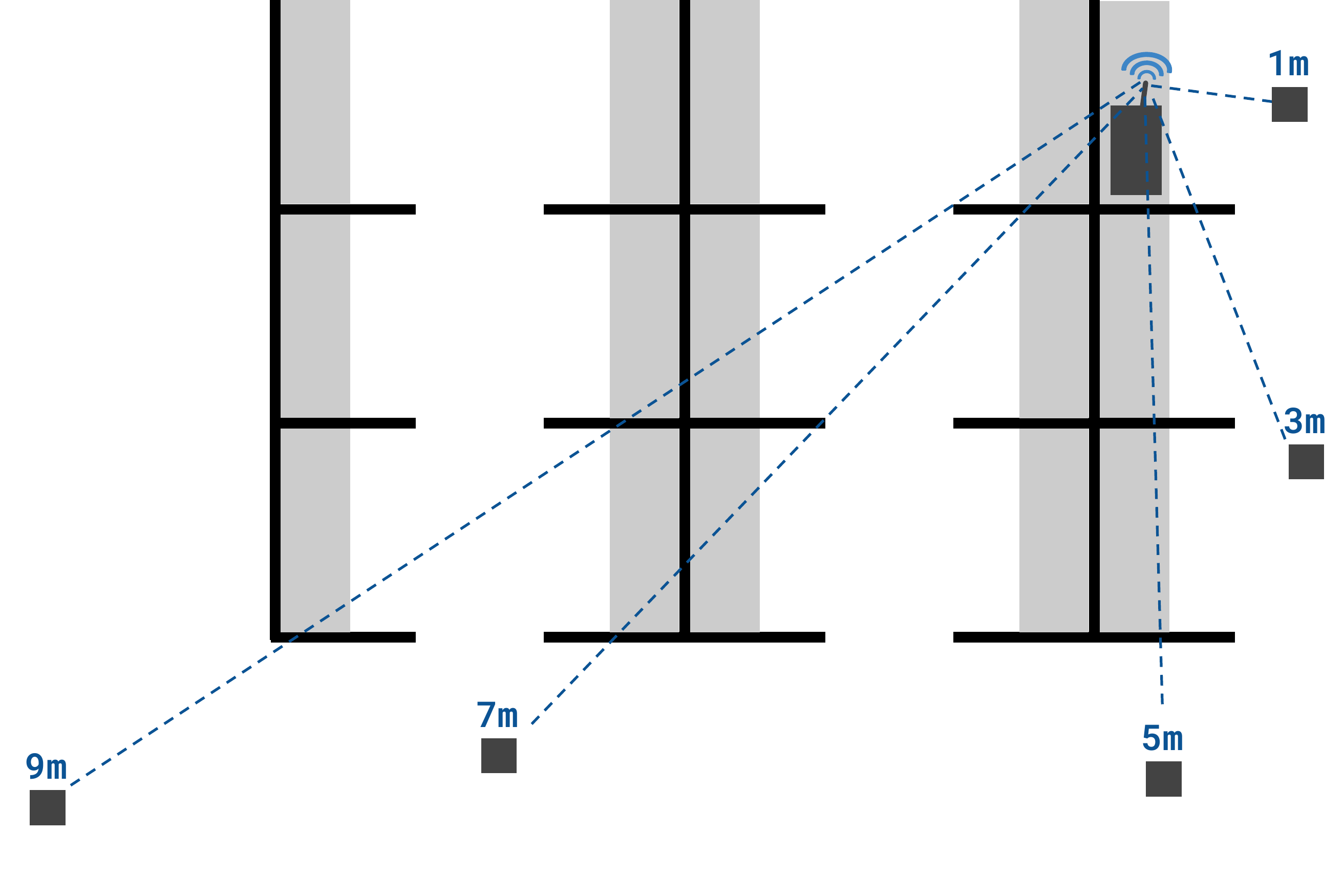}
}
\caption{\textbf{Device placements on the testing environment} (a typical office space). In the tests, we vary the distance between the edge device and the server. }
\label{fig:office}
\end{figure}

\begin{figure}
\centerline{
\includegraphics[width=\linewidth]{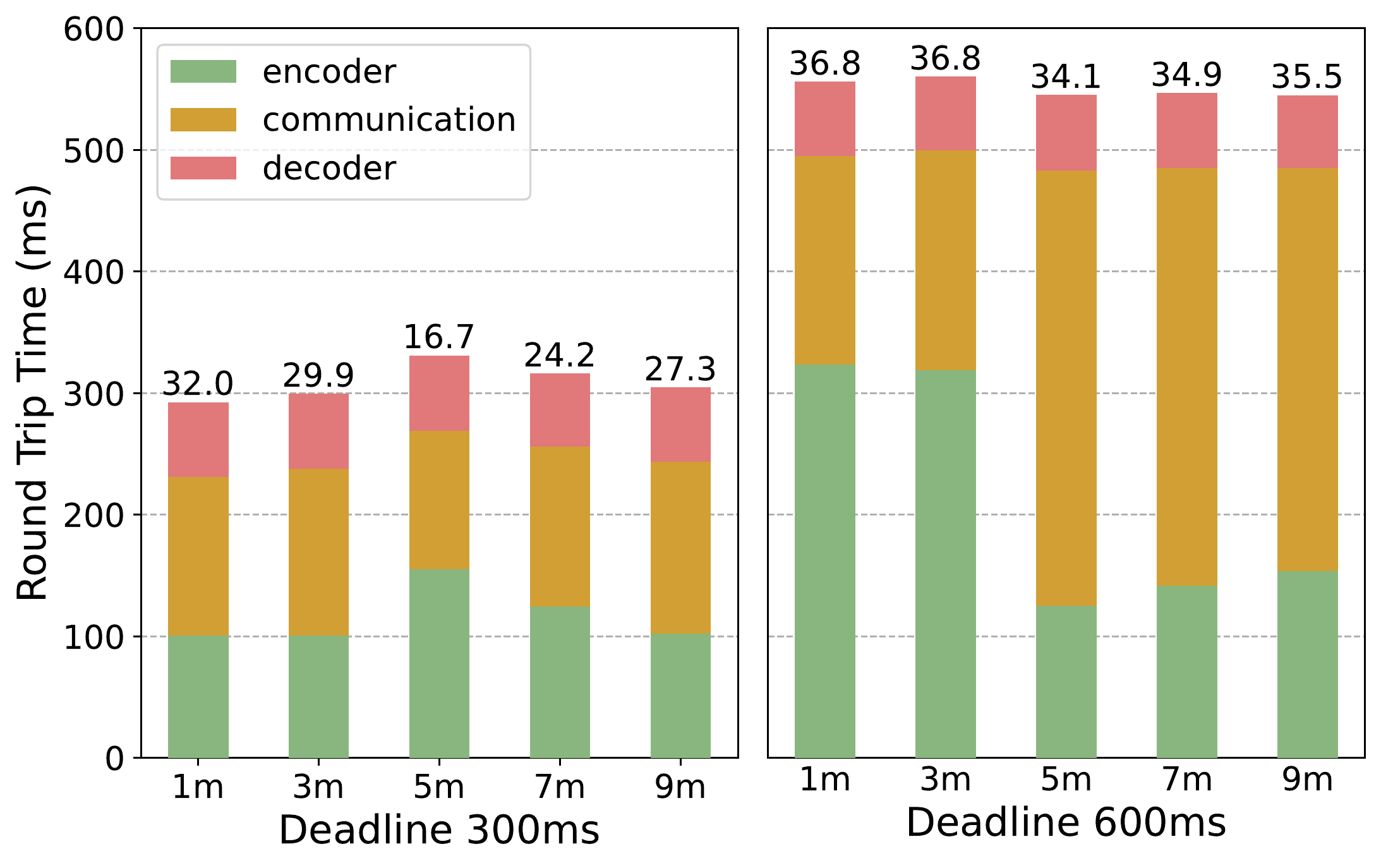}
}
\caption{ \textbf{Breakdown of RTT} delay for multiple deadlines and device placements. Numbers on top of the bars represent average \gls{map}.}
\label{fig:graph_distances}
\end{figure}

Furthermore, we evaluate a scenario corresponding to deployment adaptation, where the device is given a \gls{rtt} deadline target set by the user and at each step, a new configuration is selected based on a feedback loop. The feedback loop chooses the best possible configuration based on a table lookup containing the knowledge of the encoder execution time and the received decoder execution time provided by the server. We test our system on several transmission positions within a typical office space, graphically shown in Fig.~\ref{fig:office}. This typical use case for Bluetooth of a close-ranged application provides a predictable pattern of available capacity as a function of the distance and propagation characteristics between server and device, with bandwidth rapidly decaying in the first few meters ($<$5m), and a pattern dominated by line-of-sight availability after that. We test our dynamic system at each distance with two \gls{rtt} deadlines: 300ms and 600ms. A breakdown of latency and average \gls{map} is shown in Fig.~\ref{fig:graph_distances}. It can be noticed that the system is capable to adapt its configuration to channel variations at each placement with precision values decreasing as the distance increases. This trend does not apply to 5m, which had the worst average data rate, resulting in degraded precision. We attribute this result to unfortunate propagation characteristics of that position due to obstructions.

\begin{figure}[h]
\centerline{
\includegraphics[width=0.95\linewidth]{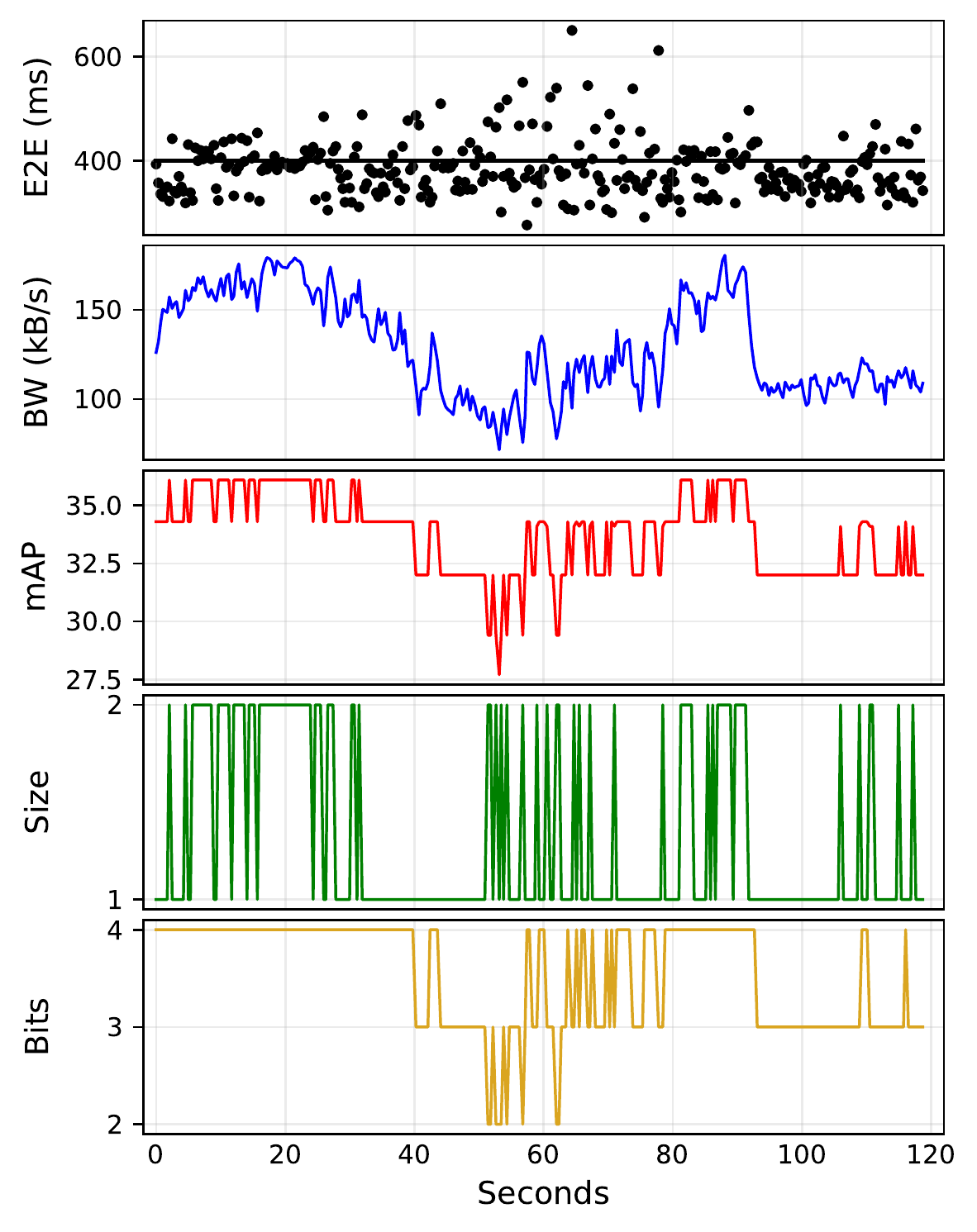}
}
\caption{\textbf{Dynamic behavior of the system.}}
\label{fig:dynamic}
\end{figure}

Finally, we show the full dynamic behavior of the system in Figure \ref{fig:dynamic}, where the user walks with the device in a range of distances between $1$ and $9$m configured for a target deadline of 400ms. From these results, we can see that most of the switching occurs on the model bandwidth, which is expected. Bandwidth is a much more stringent factor for the algorithm precision and only when bandwidth is plenty, and transmitting the full 4-bit bitrate does not meet the deadline, larger ensemble sizes are chosen by the algorithm.

\section{Conclusions}
\label{sec:concl}
In this paper, we presented a novel architecture and training for the design of encoders in the context of split computing. Different from existing solutions, the one we propose provides low-overhead low-complexity adaptation at runtime. The performance and adaptation capabilities of the proposed model are assessed by means of real-world experimentation on widely used platforms and communication devices.

\section*{Acknowledgment}

This work was supported by the Intel Corporation, Cisco and the NSF grant MLWiNS-2003237 and CCF-2140154. This research was partially conducted with support from Instituto Eldorado, which provided hardware and financial support. We extend our thanks to Eduardo Lima for arranging the lease and Alison Venancio for his technical support.

\bibliographystyle{IEEEtran}
\bibliography{bibliography}

\end{document}